\documentclass[11pt,letterpaper]{article}
\usepackage[english]{babel}
\usepackage{amsfonts}
\usepackage{graphicx}
\usepackage{amssymb}
\usepackage{amsmath}
\usepackage{fullpage}
\usepackage{booktabs}
\usepackage{longtable}
\usepackage{color}
\usepackage{authblk}
\usepackage[round]{natbib}
\usepackage{enumitem}
\usepackage{authblk}
\usepackage{fullpage}
\usepackage{amsthm}
\usepackage{multirow}
\usepackage{multicol}
\usepackage{wrapfig}
 \usepackage{url}
 \usepackage[toc,page]{appendix}

\newtheorem{theorem}{Theorem}[section]

\newtheorem{proposition}{Proposition}[section]

\newtheorem{remark}{Remark}[section]
\allowdisplaybreaks

\DeclareFontEncoding{FMS}{}{}
\DeclareFontSubstitution{FMS}{futm}{m}{n}
\DeclareFontEncoding{FMX}{}{}
\DeclareFontSubstitution{FMX}{futm}{m}{n}
\DeclareSymbolFont{fouriersymbols}{FMS}{futm}{m}{n}
\DeclareSymbolFont{fourierlargesymbols}{FMX}{futm}{m}{n}
\DeclareMathDelimiter{\VERT}{\mathord}{fouriersymbols}{152}{fourierlargesymbols}{147}

\DeclareMathOperator*{\esssup}{ess\,sup}

\begin{document}

\title{Convex Regression with a Penalty}

\author{Eunji~Lim\thanks{Corresponding author.  1 South Ave, Garden City, NY 11530; elim@adelphi.edu; +1-516-877-3811}}
\affil{Department of Decision Sciences and Marketing, Adelphi University}

\date{}
\maketitle

\begin{abstract}
A common way to estimate an unknown convex regression function $f_0: \Omega \subset \mathbb{R}^d \rightarrow \mathbb{R}$ from a set of $n$ noisy observations is to fit a convex function that  minimizes the sum of squared errors. However, this estimator is known for its tendency to overfit near the boundary of $\Omega$, posing significant challenges in real-world applications. In this paper, we introduce a new estimator of $f_0$ that avoids this overfitting by minimizing a penalty on the subgradient while enforcing an upper bound $s_n$ on the sum of squared errors. The key advantage of this method is that $s_n$ can be directly estimated from the data. We establish the uniform almost sure consistency of the proposed estimator and its subgradient over $\Omega$  as $n \rightarrow \infty$ and derive convergence rates. The effectiveness of our estimator is illustrated through its application to estimating waiting times in a single-server queue.
\end{abstract}

\noindent%


\section{Introduction}\label{sec:Intro}

Convex regression aims to estimate an unknown function $f_0:\Omega \subset \mathbb{R}^d \rightarrow \mathbb{R}$ from noisy observations $(\boldsymbol{X}_1, y_1), \cdots, (\boldsymbol{X}_n, y_n)$, under the assumption that $f_0$ is convex. We model the observations as
\[y_i = f_0(\boldsymbol{X}_i) + \varepsilon_i\]
for $i = 1, \cdots, n$, where $\Omega = [a, b]^d \subset \mathbb{R}^d$, the $\boldsymbol{X}_i$'s are independent and identically distributed (iid) $\Omega$-valued random vectors, and the error terms $\varepsilon_i$ satisfy $\mathbb{E}[\varepsilon_i] = 0$ and $\mbox{var}[\varepsilon_i] < \infty$. Given the convexity of $f_0$, a straightforward way to recover it from the data is to fit a convex function that minimizes the sum of squared errors. This leads to the following minimization problem:
\begin{equation}
\label{eqn:01} 
\begin{aligned}
&\mbox{Minimize} & & \frac{1}{n}\sum_{i = 1}^n (y_i - f(\boldsymbol{X}_i))^2
\end{aligned}
\end{equation}
over convex functions $f:\Omega \rightarrow \mathbb{R}$. Though this minimization problem appears to be infinite-dimensional, a solution can  be obtained by solving a finite-dimensional quadratic program (QP). This solution is known as the convex regression estimator, and its theoretical properties have been extensively studied in the literature; see \cite{Hildreth1954} and \cite{Kuosmanen2008} for its QP formulations; \cite{HansonPledger1976}, \cite{SeijoSen2011}, and \cite{LimGlynn2012} for almost sure (a.s.) consistency; \cite{GroeneboomEtAt2001} and \cite{HanWellner2016} for convergence rates; \cite{ChatterjeeEtAl2015} for risk bounds; \cite{HannahDunson2013} for computational algorithms; and \cite{JohnsonJiang2018} for a survey.

Convex regression is widely used in practice because it does not require the underlying function $f_0$ to have any specific parametric form; $f_0$ only needs to be convex. Many regression functions encountered in real-world applications are known or presumed to be convex. For instance, the sojourn time of a customer in a tandem queue is known to be convex as a function of the mean service times \citep{ShanthikumarYao1991}, and the cost function of some electricity distribution companies can be assumed to be convex  as a function of certain performance measures \citep{KuosmanenJohnson2020}.

\begin{figure}
\caption{The circles are the $(\boldsymbol{X}_i, y_i)$ values, the solid line is  $f_0$, and the dotted line is the convex regression estimator.}
\begin{center}
\includegraphics[scale = 0.4]{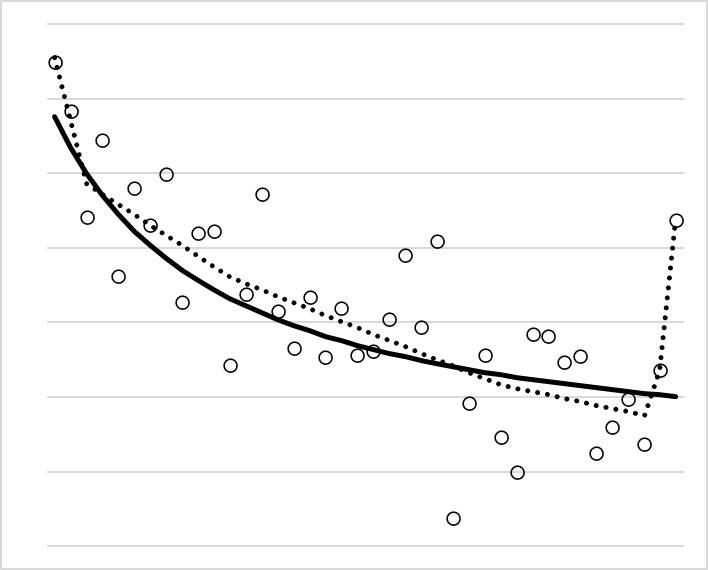}
\end{center}
\label{figure08}
\end{figure}

Despite its popularity, the convex regression estimator has a significant drawback; it tends to overfit near the boundary of its domain. This overfitting leads to inconsistent estimates and excessively large subgradients as $n$ increases near the boundary of $\Omega$; see Lemma 5.1 of \cite{GhosalSen2017}  for  theoretical evidence of overfitting and Figure \ref{figure08} for a visual representation of this issue. Such behavior poses serious challenges, particularly when estimating the subgradient of $f_0$ in these regions. For example, if $f_0$ represents the cost function associated with a company's performance metrics, its subgradient indicates the marginal costs of these measures, which provide an essential insight for practitioners. Since practitioners often use the subgradient of the convex regression estimator as an estimator of 
$f_0$’s subgradient, the tendency of the estimator to produce excessively large subgradients near the boundary of $\Omega$
 creates serious practical challenges.

To address this issue of overfitting, various methods have been proposed. One approach is to limit the magnitude of the subgradients by adding a penalty term to the objective function of  (\ref{eqn:01}). Specifically, we can solve
\begin{equation*}
\begin{aligned}
& \mbox{(A):}& &  \min_{f \in \mathcal{F}} &  & \frac{1}{n}\sum_{i = 1}^n (y_i - f(\boldsymbol{X}_i))^2 + \lambda_n J(f),
\end{aligned}
\end{equation*}
where $\lambda_n$ is a nonnegative constant called the smoothing constant, $J(f)$ is a penalty term measuring the overall magnitude of the subgradient of $f$, and
\[\mathcal{F} \triangleq \{f:\Omega\rightarrow \mathbb{R} \mid f \mbox{ is convex and }J(f) < \infty\}.\]
The solution to Problem (A) is referred as the penalized convex regression estimator.

Problem (A) suggests a way to bound the magnitude of the subgradient, but its numerical performance is highly sensitive to the selection of $\lambda_n$. A common method for determining $\lambda_n$ is cross validation, but this method has significant limitations when applied in numerical contexts. Cross-validation is a heuristic that minimizes a cross-validation function, which is a non-convex function defined over a set of nonnegative real numbers. This non-convex optimization often requires a time-consuming trial-and-error process. Moreover, cross validation yields $\lambda_n = 0$ a few percent of the time, reverting Problem (A) back to (\ref{eqn:01}); see page 65 of \cite{Wahba1990}. Given that (\ref{eqn:01}) produces inconsistent estimates of $f_0$ and excessively large subgradients near the boundary of its domain, this creates a considerable challenge. 

Another approach to constraining the subgradient of the convex regression estimator is enforcing a bound on the subgradients by adding a constraint directly to (\ref{eqn:01}) as follows:
\begin{equation*}
\begin{aligned}
   & \mbox{(B):}& &  \min_{f \in \mathcal{F}} & &  \frac{1}{n}\sum_{i = 1}^n (y_i - f(\boldsymbol{X}_i))^2   \\
   & & &  \mbox{subject to}\quad && J(f) \leq u_n
\end{aligned}
\end{equation*}
for some nonnegative constant $u_n$. One downside of this approach is that its numerical performance is highly sensitive to the selection of $u_n$, yet there is little guidance on how to choose $u_n$ beyond using cross validation. 

To address these practical challenges, we propose a new method for estimating a convex function $f_0$. Rather than minimizing the sum of squares while imposing a bound on the penalty, we minimize the penalty while limiting the sum of squared errors as follows:
\begin{equation*}
\begin{aligned}
    & \mbox{(C):}&  &  \min_{f \in \mathcal{F}} & &  J(f)  \\
    & & & \mbox{subject to}\quad& &\frac{1}{n}\sum_{i = 1}^n (y_i - f(\boldsymbol{X}_i))^2  \leq s_n
\end{aligned}
\end{equation*}
for some nonnegative constant $s_n$. The primary advantage of Problem (C) is the ease of estimating $s_n$ from the data.

The main role that $\frac{1}{n}\sum_{i = 1}^n (y_i - f(\boldsymbol{X}_i))^2  \leq s_n$ plays in Problem (C) is ensuring that $f_0$ becomes a feasible solution to Problem (C). Ideally, $s_n$ should be the smallest value that guarantees $\frac{1}{n} \sum_{i = 1}^n (y_i - f_0(\boldsymbol{X}_i))^2 \leq s_n$
because a larger  $s_n$ expands the feasible solution set for Problem (C), complicating the search for a solution. Thus, a candidate for $s_n$ is $ \frac{1}{n}\sum_{i = 1}^n (y_i - f_0(\boldsymbol{X}_i))^2$, which approaches $\mathbb{E}[\mbox{var}(\varepsilon_1|\boldsymbol{X}_1)]$ as $n \rightarrow \infty$ when $\mbox{var} [\varepsilon_i]$ depends only on $\boldsymbol{X}_i$ and the $\varepsilon_i$'s are independent given $((\boldsymbol{X}_i, y_i) : i \geq 1)$. 

There are several methods for estimating $\mathbb{E}[\mbox{var}(\varepsilon_1|\boldsymbol{X}_1)]$  from the data set  $ ((\boldsymbol{X}_i, y_i) : i = 1, \cdots, n)$, and we will outline two of them here. 
\begin{enumerate}
\item[(1)] We can partition $[a, b]^d$ into $r$ hyperrectangles, denoted by $H_1, \cdots, H_r$, compute the sample variance $V_j$ of the $y_i$'s within each $H_j$ for $j \in \{1, \cdots, r\}$, and then estimate $s_n$ using \[\mathbb{P}(\boldsymbol{X}_1 \in H_1)\cdot V_1 + \cdots + \mathbb{P} (\boldsymbol{X}_1 \in H_r) \cdot V_r.\] 
     This will be a valid estimate of $\mathbb{E}[\mbox{var}(\varepsilon_1|\boldsymbol{X}_1)]$  if the number of $\boldsymbol{X}_i$'s in each hyperrectangle increases to infinity and the volume of each hyperrectangle decreases to zero as $n \rightarrow \infty$.
     \item[(2)]In some simulation contexts, it is possible to generate multiple replications of $y_i$ at each $\boldsymbol{X}_i$. In such a case, $\mathbb{E}[\mbox{var}(\varepsilon_1|\boldsymbol{X}_1)]$ can be estimated more directly by taking multiple observations of $y_i$ at each $\boldsymbol{X}_i$, calculating the sample variance $W_i$ of these observations at $\boldsymbol{X}_i$, and then using  \[\frac{\tau(\boldsymbol{X}_1) \cdot W_1 + \cdots + \tau(\boldsymbol{X}_n) \cdot W_n}{ \tau(\boldsymbol{X}_1) + \cdots + \tau(\boldsymbol{X}_n)}\]
 as an estimate of $s_n$, where $\tau(\cdot)$ is the density function of $\boldsymbol{X}_1$. 
 \end{enumerate}
 These approaches offer practical guidance for selecting $s_n$ directly from the data, which is a significant benefit in real-world applications.

From a modeling standpoint, one of the key decisions to make is how to specify $J(f)$. There are various forms for the penalty term $J(f)$, with one of the most commonly used being the Sobolev seminorm $J_p(f)$ defined as follows:
\begin{equation*}
J_p(f) = \left\{
\begin{array}{l l}
\displaystyle\left(\left\|\frac{\partial f}{\partial x_1}\right\|_p^p + \cdots +  \left\|\frac{\partial f}{\partial x_d}\right\|^p\right)^{1/p}, & \mbox{if } 1 \leq p < \infty\\
\displaystyle\max_{1 \leq i \leq d} \left\|\frac{\partial f}{\partial x_i}\right\|_{\infty}, & \mbox{if } p = \infty,
\end{array}
\right. 
\end{equation*}
where
\[
\left\| g\right\|_p \triangleq \left\{
\begin{array}{l l}
\displaystyle\left( \int_{\Omega} |g(\boldsymbol{x})|^p d\boldsymbol{x}\right)^{1/p}, & \mbox{if } 1 \leq p< \infty\\
\displaystyle\esssup_{\boldsymbol{x} \in \Omega} |g(\boldsymbol{x})|, & \mbox{if }p = \infty.
\end{array}
\right. 
\]
Here, $\partial f/\partial x_i$ denotes the $i$th partial derivative of $f$ in the weak sense for $i = 1, \cdots, d$. The partial derivative in the weak sense is a generalized form of the classical partial derivative, such that whenever the classical partial derivative exists, it also corresponds to the partial derivative in the weak sense; see \cite{AdamsFournier2003} for a definition of partial derivatives in the weak sense.

In this paper, we focus on the case where
\[\displaystyle J(f) = J_{\infty}(f) = \max_{1 \leq i \leq d} \esssup_{\boldsymbol{x} \in \Omega}\left|\frac{\partial f}{\partial x_i} (\boldsymbol{x})\right|\]  and establish theoretical guarantees for the solutions to Problems (A) and (C). Given that our interest lies in the overfitting behavior of the convex regression estimator near the boundary of its domain, the most desirable behavior of our estimator is the uniform a.s.\ consistency over the entire domain $\Omega$ rather than pointwise a.s.\ consistency. In this paper, we establish that the solution to Problem (C) converges to $f_0$ a.s.\  uniformly over $\Omega$ as $n \rightarrow \infty$ and derive its convergence rates. Additionally, we prove that the subgradient of the solution to Problem (C) converges to the gradient of $f_0$ a.s.\ uniformly over $\Omega$  as $n \rightarrow \infty$. To our knowledge, this is the first paper to introduce Problem (C) within the context of convex regression and establish its theoretical properties. 

We also establish the theoretical properties of the solution to Problem (A). Although the theoretical properties of the solution to Problem (B) are well-established, such as the convergence rates computed by \cite{Lim2014}, \cite{BalazsEtAl2015} and \cite{MazumderEtAl2019}, there is limited understanding of the theoretical properties of the solution to Problem (A). In this paper, we establish the a.s.\ uniform consistency of the solution to Problem (A) across $\Omega$ as $n \rightarrow \infty$ and compute its convergence rates, offering theoretical assurance for using Problem (A). We also show that the subgradient of the solution to Problem (A) converges uniformly to the gradient of $f_0$ a.s.\ over $\Omega$ as $n \rightarrow \infty$.

\paragraph{Literature Review.}  Since its introduction by \cite{Hildreth1954}, convex regression has attracted significant interest from the statistics, operations research, and economics communities; see, for example, \cite{HansonPledger1976}, \cite{GroeneboomEtAt2001}, \cite{Kuosmanen2008},  \cite{KuosmanenJohnson2010}, \cite{SeijoSen2011}, \cite{LimGlynn2012}, \cite{HannahDunson2013}, \cite{Lim2014}, \cite{HanWellner2016},  \cite{Keshvari2018}, \cite{Lim2020}, \cite{ChenEtAl2021}, and \cite{EckmanEtAl2022} for theories and applications of convex regression. 

While convex regression and its applications have been a well-studied topic for decades, studies on overfitting have only recently started to gain attention. Recently, the overfitting behavior of the convex regression estimator has been observed by, for example, \cite{Balabdaoui2007}, Section 4.3 of \cite{BalazsEtAl2015}, and Lemma 5.1 of \cite{GhosalSen2017}. To mitigate overfitting, Problem (B) was initially introduced and examined by the research community; see, for example, \cite{Lim2014}, \cite{BalazsEtAl2015} and \cite{MazumderEtAl2019}. Subsequently, Problem (A) was studied by many researchers; see \cite{BlanchetEtAl2019}, \cite{BertsimasMundru2020}, \cite{ChenEtAl2020}, and \cite{ChenMazumder2024} for example. \cite{BertsimasMundru2020}, \cite{ChenEtAl2020}, and \cite{ChenMazumder2024}  developed efficient computation algorithms for solving Problem (A) when $J(f) = \frac{1}{n}\sum_{i = 1}^n |\mbox{subgrad }f(\boldsymbol{X}_i)|_2^2$ for $f \in \mathcal{F}$, where $\mbox{subgrad }f(\boldsymbol{X}_i)$ denotes the subgradient of $f$ at $\boldsymbol{X}_i$ and $|\cdot|_2$ is the Euclidean norm. \cite{BlanchetEtAl2019} computed the convergence rates when $d > 4$, for the case where  $\frac{1}{n}\sum_{i = 1}^n |y_i - f(\boldsymbol{X}_i)| + \lambda_n J_{\infty}(f)$ is minimized over a set of convex and Lipschitz functions. Our findings on Problem (A) extend their work by computing the convergence rate of the solution to Problem (A) for all values of $d$ including the case where $d \leq 4$. Moreover, our convergence rates for Problem (A), when $d > 4$, are a slight improvement over those obtained by \cite{BlanchetEtAl2019} by a logarithmic factor. Additionally, we establish the uniform a.s.\ consistency of the solution to Problem (A) and its subgradient, a topic that has not been previously investigated. Problem (C) has not been introduced in the context of convex regression, and it is the new framework we present in this paper.

This paper can also be viewed as a contribution to the
larger literature on shape-restricted regression. One of the earliest studies in this field is isotonic regression, which assumes the underlying function $f_0$ is monotone; see, for example, \cite{Brunk1958}, \cite{BarlowBrunk1972}, and \cite{Wright1979}. Within the isotonic regression framework, \cite{Pal2008} and \cite{WuEtAl2015} proposed a penalized isotonic estimator, which is a version of Problem (A) for monotone functions.  \cite{LussRosset2017} introduced a bounded isotonic regression estimator, which is a version of Problem (B) for monotone functions. However, no estimator that corresponds to Problem (C) for monotone functions has been proposed so far. As a result, applying Problem (C) to monotone regression is still an open question.

\paragraph{Main Contributions of This Paper.} The main contributions of this paper can be summarized  as follows.
\begin{enumerate}
\item[(1)] Introduction of Problem (C):  We tackle the overfitting issues commonly encountered in convex regression estimators and introduce a new method, referred to as Problem (C), that is designed to mitigate overfitting. This method has a significant practical advantage over existing methods, Problems (A) and (B), because its parameter can be easily estimated from the data set.

\item[(2)] Theoretical Guarantees for Problem (C): The paper establishes that the solution to Problem (C) and its subgradient converge to $f_0$ and $\nabla f_0$, respectively, a.s.\ uniformly over the entire domain $\Omega$
as $n \rightarrow \infty$. We also derive the convergence rates for the solution to Problem (C).  This contrasts with traditional methods that only ensure consistency at the interior of $\Omega$.  Our findings provide a theoretical basis for using Problem (C) to mitigate overfitting. 

\item[(3)] Theoretical Guarantees for Problem (A):  We prove that the solution to Problem (A) and its subgradient converge to $f_0$ and $\nabla f_0$, respectively, a.s.\ uniformly over the entire domain $\Omega$
as $n \rightarrow \infty$. Additionally, we obtain the convergence rates for the solution to Problem (A). These were previously open questions that have not been resolved until now.
\end{enumerate}

\paragraph{Organization of This Paper.} This paper is organized as follows. Section \ref{QP_formulation} outlines the mathematical framework for our analysis and demonstrates that the solutions to Problems (A) and (C) can be obtained by solving a QP and a quadratically constrained quadratic program (QCQP), respectively. In Section \ref{label:Asymptotic_Properties}, we establish the theoretical properties of the solutions to Problems (A) and (C).  Section \ref{label:numerical_results} illustrates their numerical performance. Finally, Section \ref{label:conclusions} provides some concluding remarks.

\paragraph{Notation and Definitions} Throughout this paper, we view vectors as columns and write $\boldsymbol{x}^{\mathsf{T}}$ to denote the transpose of a vector $\boldsymbol{x} \in \mathbb{R}^d$. For $\boldsymbol{x} \in \mathbb{R}^d$, we write its $i$th component as $x_i$, so $\boldsymbol{x} = (x_1, \cdots, x_d)^{\mathsf{T}}$. We define $|\boldsymbol{x}|_p = (|x_1|^p + \cdots + |x_d|^p)^{1/p}$ for $p = 1, 2, \cdots$ and $|\boldsymbol{x}|_{\infty} = \max_{1 \leq i 
\leq d}|x_i|$.

For any convex set $S \subset \mathbb{R}^d$ and a convex function $g:S \rightarrow \mathbb{R}$, a vector $\boldsymbol{\beta} \in \mathbb{R}^d$ is said to be a subgradient of $g$ at $\boldsymbol{x} \in S$ if $g(\boldsymbol{y}) \geq g(\boldsymbol{x}) + \boldsymbol{\beta}^{\mathsf{T}}(\boldsymbol{y} - \boldsymbol{x})$ for all $\boldsymbol{y} \in S$. We denote a subgradient of $f$ at $\boldsymbol{x} \in S$ by $\mbox{subgrad }g(\boldsymbol{x})$. The set of all subgradients of $g$ at $\boldsymbol{x}$ is called the subdifferential of $g$ at $\boldsymbol{x}$, denoted by $\partial g(\boldsymbol{x})$. The subdifferential $\partial g(\boldsymbol{x})$ of a convex function $g:S \rightarrow \mathbb{R}$ is non--empty for any $\boldsymbol{x} \in S$; see Theorem 23.4 on page 217 of \cite{Rockafella}. When a function $g:S \rightarrow \mathbb{R}$ is differentiable at $\boldsymbol{x} \in [a, b]^d$, we denote its gradient at $\boldsymbol{x}$ by $\nabla g(\boldsymbol{x})$.

Let $(a_n: n \geq 1)$ and $(b_n: n \geq 1)$ be sequences of positive real numbers. We say $a_n = O(b_n)$ if there exist positive real numbers $c$ and $n_0$ such that $a_n \leq c b_n$ for all $n \geq n_0$.

For a sequence of random variables $(Z_n : n \geq 1)$ and a sequence of positive real numbers $(\alpha_n :  n \geq 1)$, we say $Z_n = \mathcal{O}_p(\alpha_n)$ as $n \rightarrow \infty$ if, for any $\epsilon > 0$, there exist constants $c$ and $n_0$ such that $\mathbb{P}(|Z_n /\alpha_n| > c) < \epsilon$ for all $n \geq n_0$.

Let $\mathcal{C}$ be a class of functions $g:\Gamma \rightarrow \mathbb{R}$ for $\Gamma \subset \mathbb{R}^d$. For $u > 0$, let $N(u, \mathcal{C})$ be the smallest value of $N$ such that there exists $\{g_1, \cdots, g_N\}$ satisfying the following condition: for any $g \in \mathcal{C}$, there exists $j \in \{1, \cdots, N\}$ such that $\sup_{ \boldsymbol{x} \in \Gamma} |g(\boldsymbol{x}) - g_j(\boldsymbol{x})| \leq u$. If no such set exists, we let $N(u, \mathcal{C}) = \infty$. We refer to $N(u, \mathcal{C})$ as the $u$-covering number in the supremum norm. We also call $H(u, \mathcal{C}) = \log(N(u, \mathcal{C}) + 1)$ the $u$-entropy of $\mathcal{C}$ in the supremum norm.

\section{QP Formulations of Problems (A), (B), and (C)}
\label{QP_formulation}

Throughout this paper, we focus on the case where $J(f) = J_{\infty} (f)$. This section is concerned with the QP formulations of Problems (A), (B) and (C).  Propositions \ref{proposition:01}, \ref{proposition:02}, and \ref{proposition:03} establish the QP formulations of Problems (A), (B), and (C), respectively. Propositions \ref{proposition:01} and \ref{proposition:02} are our new results while Proposition \ref{proposition:02} is a previously established result included here for completeness.

It should be noted that for any  $f \in \mathcal{F}$,
\begin{equation*}
J_{\infty} (f)  \triangleq  \max_{1 \leq i \leq d} \esssup_{\boldsymbol{x} \in \Omega}\left|\frac{\partial f}{\partial x_i} (\boldsymbol{x})\right| =  \max_{1 \leq i \leq d} \sup_{\boldsymbol{x} \in \Omega'}\left|\frac{\partial f}{\partial x_i} (\boldsymbol{x})\right|
\end{equation*}
for any set $\Omega'$ of measure one. Because a convex function over $\Omega$ is differentiable almost everywhere, as in  Theorem 25.5 on page 246 of \cite{Rockafella}, the set $\Omega_f \triangleq \{\boldsymbol{x} \in \Omega \mid f \mbox{ is differentiable at } \boldsymbol{x}\}$ has measure one. Thus,
 \begin{equation}
\label{eqn:02}
J_{\infty} (f) = \sup_{x \in \Omega_f} |\nabla f  (\boldsymbol{x})|_{\infty}
\end{equation}
for any  $f \in \mathcal{F}$.

Proposition \ref{proposition:01} states that a solution to Problem (A) can be obtained by solving a QP when $J(f) = J_{\infty}(f)$. The proof of Proposition \ref{proposition:01} is provided in Appendix \ref{sec:appendixA}.

\begin{proposition}
\label{proposition:01} 
Let $\lambda_n \geq 0$ be given. Consider the following QP in the decision variables $f_1, \cdots, f_n \in \mathbb{R}$, $\boldsymbol{\beta}_1, \cdots, \boldsymbol{\beta}_n \in \mathbb{R}^d$, and $M \in \mathbb{R}$:
\begin{equation}
\label{ProblemA_QP}
\begin{aligned}
& \mbox{min} &  &\frac{1}{n} \sum_{i = 1}^n (y_i - f_i)^2 + \lambda_n M &&  \\
&\mbox{subject to} &  &f_i \geq f_j + \boldsymbol{\beta}^{\mathsf{T}}_j(\boldsymbol{X}_i - \boldsymbol{x}_j) &  & 1\leq i, j \leq n\\
& & &|\boldsymbol{\beta}_i|_{\infty} \leq M &  &  1\leq i \leq n.
\end{aligned}
\end{equation}
Then, (\ref{ProblemA_QP}) has a minimizer $\hat{f}_1, \cdots, \hat{f}_n, \hat{\boldsymbol{\beta}}_1, \cdots, \hat{\boldsymbol{\beta}}_n, \hat{M}$. Furthermore, $\hat{f}_1, \cdots, \hat{f}_n, \hat{M}$ are unique. 

Let $\hat{f}_n^*:\Omega \rightarrow \mathbb{R}$ be defined by
\begin{equation}
\label{eqn:03}
\hat{f}_n^*(\boldsymbol{x}) = \max\{\hat{f}_i + \hat{\boldsymbol{\beta}}_i^{\mathsf{T}} (\boldsymbol{x} - \boldsymbol{X}_i)\}
\end{equation}
for $\boldsymbol{x} \in \Omega$. Then, $\hat{f}_n^*$ is a solution to Problem (A) with $J(f) = J_{\infty}(f)$.
\end{proposition}

Proposition \ref{proposition:02} presents a QP formulation for Problem (B) when $J(f) = J_{\infty}(f)$. The proof of Proposition \ref{proposition:02}  can be found in, for example,  Proposition 1 of \cite{Lim2014} and Theorem 1 of \cite{LinEtAl2022}. 

\begin{proposition}
\label{proposition:02} 
Let $(u_n:  n \geq 1)$ be a sequence of non-negative real numbers. Consider the following QP in the decision variables $f_1, \cdots, f_n \in \mathbb{R}$, $\boldsymbol{\beta}_1, \cdots, \boldsymbol{\beta}_n \in \mathbb{R}^d$:
\begin{equation}
\label{ProblemB_QP}
\begin{aligned}
& \mbox{min} &  &\frac{1}{n} \sum_{i = 1}^n (y_i - f_i)^2  &&  \\
&\mbox{subject to} &  &f_i \geq f_j + \boldsymbol{\beta}^{\mathsf{T}}_j(\boldsymbol{X}_i - \boldsymbol{x}_j) &  &  1\leq i, j \leq n\\
& & &|\boldsymbol{\beta}_i|_{\infty} \leq u_n &  &  1\leq i \leq n.
\end{aligned}
\end{equation}
Then, (\ref{ProblemB_QP}) has a minimizer $\hat{f}, \cdots, \hat{f}_n, \hat{\boldsymbol{\beta}}_1, \cdots, \hat{\boldsymbol{\beta}}_n$. Furthermore, the values $\hat{f}_1, \cdots, \hat{f}_n$ are unique. 

Let $\hat{f}_n^{**}:\Omega \rightarrow \mathbb{R}$ be defined by
\begin{equation}
\label{eqn:04}
\hat{f}_n^{**}(\boldsymbol{x}) = \max\{\hat{f}_i + \hat{\boldsymbol{\beta}}_i^{\mathsf{T}} (\boldsymbol{x} - \boldsymbol{X}_i)\}
\end{equation}
for $\boldsymbol{x} \in \Omega$. Then, $\hat{f}_n^{**}$ is a solution to Problem (B) with $J(f) = J_{\infty}(f)$.
\end{proposition}

Our next proposition, Proposition \ref{proposition:03}, asserts that a solution to Problem (C) can be obtained by solving a QCQP. The proof of Proposition \ref{proposition:03} is provided in Appendix \ref{sec:appendixB}.

\begin{proposition}
\label{proposition:03} 
Assume that given $(\boldsymbol{X}_i: i \geq 1)$, the $\varepsilon_i$'s are iid  and that the $\varepsilon_i$'s are uniformly sub-Gaussian, i.e., 
there exist positive constants $K$ and $\sigma_0$ such that
\[ \max_{1 \leq i \leq n} K^2 (\mathbb{E}[\exp (\varepsilon_i^2/K^2)] - 1) \leq \sigma_0^2.\]
In addition, suppose $s_n  = \sigma^2 + c n^{-1/2}\sqrt{\log n}$ for some constant $c > 8(K^2 + \sigma_0^2)$. 

Then, $f_0$ satisfies 
\[\frac{1}{n}\sum_{i = 1}^n (y_i - f_0(\boldsymbol{X}_i))^2 \leq s_n\]
for all $n$ sufficiently large a.s.

Consider the following QCQP in the decision variables $f_1, \cdots, f_n \in \mathbb{R}$, $\boldsymbol{\beta}_1, \cdots, \boldsymbol{\beta}_n \in \mathbb{R}^d$, and $M \in \mathbb{R}$:
\begin{equation}
\label{ProblemC_QP}
\begin{aligned}
& \mbox{min} &  &M &&  \\
&\mbox{subject to} &  &f_i \geq f_j + \boldsymbol{\beta}^{\mathsf{T}}_j(\boldsymbol{X}_i - \boldsymbol{x}_j) &  &  1\leq i, j \leq n\\
& & &|\boldsymbol{\beta}_i|_{\infty} \leq M &  & 1\leq i \leq n.\\
& & &\frac{1}{n}\sum_{i= 1}^n (y_i - f_i)^2 \leq s_n &  & 1\leq i\leq n.
\end{aligned}
\end{equation}
Then, (\ref{ProblemC_QP}) has a minimizer $\hat{f}_1, \cdots, \hat{f}_n, \hat{\boldsymbol{\beta}}_1, \cdots, \hat{\boldsymbol{\beta}}_n$, $\hat{M}$ for $n$ sufficiently large a.s., and the value $\hat{M}$ is unique. 
 
Let $\hat{f}_n^{***}:\Omega \rightarrow \mathbb{R}$ be defined by
\begin{equation}
\label{eqn:05}
\hat{f}_n^{***}(\boldsymbol{x}) = \max\{\hat{f}_i + \hat{\boldsymbol{\beta}}_i^{\mathsf{T}} (\boldsymbol{x} - \boldsymbol{X}_i)\}
\end{equation}
for $\boldsymbol{x} \in \Omega$. Then, $\hat{f}_n^{***}$ is a solution to Problem (C) with $J(f) = J_{\infty}(f)$.
\end{proposition}

\section{Main Results}
\label{label:Asymptotic_Properties}

We are now ready to present our main results, Theorems \ref{thm:A} and \ref{thm:C}. Theorem \ref{thm:B} is a previously established result included here for completeness.
 
To analyze $\hat{f}_n^*$, $\hat{f}_n^{**}$, and $\hat{f}_n^{***}$, solutions to Problems (A), (B), and (C), we need the following assumptions.
\begin{enumerate}

\item[A1.] $\Omega = [a, b]^d$, $J(f) = J_{\infty}(f)$ for $f \in \mathcal{F}$, and $f_0 \in \mathcal{F}$.

\item[A2.] $(\boldsymbol{X}_1, y_1), (\boldsymbol{X}_2, y_2), \cdots$ is a sequence of iid, $\Omega \times \mathbb{R}$-valued random vectors satisfying 
    \[ \quad y_i = f_0(\boldsymbol{X}_i) + \varepsilon_i \quad \mbox{ for }\quad i \geq 1. \]

\item[A3.] For any subset $A \subset \Omega$ with nonempty interior, $\mathbb{P}(\boldsymbol{X}_1 \in A) > 0$.
\item[A4.] Given $(\boldsymbol{X}_i: i \geq 1)$, the $\varepsilon_i$'s are iid and satisfy $\mathbb{E}[\varepsilon_i] = 0$ and $\mbox{var}[\varepsilon_i] = \sigma^2$ for $i \geq 1$.
\item[A5.] The $\varepsilon_i$'s are uniformly sub-Gaussian, i.e., there exist positive constants $K$ and $\sigma_0$ such that
\[ \max_{1 \leq i \leq n} K^2 (\mathbb{E}[\exp (\varepsilon_i^2/K^2)] - 1) \leq \sigma_0^2.\]
\item[A6.] $f_0$ is differentiable on $\Omega$.
\end{enumerate}

Our first theorem, Theorem \ref{thm:A}, establishes that $\hat{f}_n^*$ and its subgradient converge to $f_0$ and $\nabla f_0$, respectively, both uniformly over $\Omega$ as $n \rightarrow \infty$ with probability one.   We also obtain convergence rates. The proof of Theorem \ref{thm:A} is provided in Appendix \ref{sec:appendixC}.
\begin{theorem} 
\label{thm:A}
Let \[\delta_n^2 = \left\{
\begin{array}{l l}
n^{-\frac{4}{4 + d}}, & \mbox{ if } d = 1, 2, 3\\
n^{-1/2} \log n, & \mbox{ if } d = 4\\
n^{-2/d}, & \mbox{ if } d \geq 5
\end{array}
\right.\]
for $n \geq 1$. In addition, let $(\lambda_n : n \geq 1)$ be a sequence of nonnegative real numbers satisfying (i) $\lambda_n \rightarrow 0$ as $n \rightarrow \infty$ and (ii) $\delta_n^2 = O(\lambda_n)$.

\begin{enumerate}
\item[(a)]
Under A1--A5,  we have
\[\sup_{\boldsymbol{x} \in \Omega} |\hat{f}_n^* (\boldsymbol{x}) - f_0(\boldsymbol{x})| \rightarrow 0\]
as $n \rightarrow \infty$ with probability one.
\item[(b)]
Under A1--A6, we have 
\[\sup_{\boldsymbol{x} \in \Omega} \sup_{\boldsymbol{\beta} \in \partial \hat{f}_n^* (\boldsymbol{x})} |\boldsymbol{\beta} - \nabla f_0 (\boldsymbol{x})|_{\infty} \rightarrow 0\]
as $n \rightarrow \infty$ with probability one.
\item[(c)]
Assume $\lambda_n = O(\delta_n^2)$. Under A1--A5,  we have 
\begin{equation*}
\frac{1}{n}\sum_{i = 1}^n (\hat{f}_n^{*}(\boldsymbol{X}_i) - f_0(\boldsymbol{X}_i))^2 = \left\{
\begin{array}{l l}
\mathcal{O}_p(n^{-\frac{4}{4 + d}}), & \mbox{ if } d = 1, 2, 3\\
\mathcal{O}_p(n^{-1/2} \log n), & \mbox{ if } d = 4\\
\mathcal{O}_p(n^{-2/d}), & \mbox{ if } d \geq 5
\end{array}
\right.
\end{equation*}
as $n \rightarrow \infty$.
\end{enumerate} 
\end{theorem}

\begin{remark}
\label{rmk:01} It should be noted that \cite{BlanchetEtAl2019} presented results comparable to Theorem 4.1 (c) for the case where $d \geq 5$. In fact, our results are a slight improvement over their results by a logarithmic factor when $d > 5$. Also, we could not find any prior research on the convergence rates for the case where $d = 1, 2, 3,$ or $4$. Additionally,  to the best of our knowledge, no existing work has established the a.s.\ uniform consistency of $\hat{f}_n^*$ and its subgradient.
\end{remark}

We next turn to Theorem \ref{thm:B} that  describes the asymptotic behavior of $\hat{f}_n^{**}$. Theorem \ref{thm:B} states that $\hat{f}_n^{**}$ and its subgradient converge to $f_0$ and $\nabla f_0$, respectively, both uniformly over $\Omega$ as $n \rightarrow \infty$ with probability one. Additionally, it provides the convergence rate for $\hat{f}_n^{**}$. For the proofs of Theorem \ref{thm:B} (a) and (b), refer to Theorem 4 of \cite{LiaoEtAl2024}, and for the proof of Theorem \ref{thm:B} (c), see Theorem 2 of Theorem 1 of \cite{Lim2014}.

\begin{theorem}
\label{thm:B} Let $(u_n : n \geq 1)$ be a sequence of nonnegative real numbers, satisfying $J_{\infty} (f_0) \leq u_n$ for all $n \geq 1$. 

\begin{enumerate}
\item[(a)]
Under A1--A5,  we have 
\[\sup_{\boldsymbol{x} \in \Omega} |\hat{f}_n^{**} (\boldsymbol{x}) - f_0(\boldsymbol{x})| \rightarrow 0\]
as $n \rightarrow \infty$ with probability one.
\item[(b)]
Under A1--A6, we have 
\[\sup_{\boldsymbol{x} \in \Omega} \sup_{\boldsymbol{\beta} \in \partial \hat{f}_n^{**} (\boldsymbol{x})} |\boldsymbol{\beta} - \nabla f_0(\boldsymbol{x})| \rightarrow 0\]
as $n \rightarrow \infty$ with probability one.
\item[(c)]
Under A1--A5, we have 
\begin{equation*}
\frac{1}{n}\sum_{i = 1}^n (\hat{f}_n^{**}(\boldsymbol{X}_i) - f_0(\boldsymbol{X}_i))^2 = \left\{
\begin{array}{l l}
\mathcal{O}_p(n^{-\frac{4}{4 + d}}), & \mbox{ if } d = 1, 2, 3\\
\mathcal{O}_p(n^{-1/2} \log n), & \mbox{ if } d = 4\\
\mathcal{O}_p(n^{-2/d}), & \mbox{ if } d \geq 5
\end{array}
\right.
\end{equation*}
as $n\rightarrow \infty$.
\end{enumerate}

\end{theorem}

Our next theorem, Theorem \ref{thm:C}, establishes that $\hat{f}_n^{***}$ and its subgradient converge to $f_0$ and $\nabla f$, respectively, both uniformly over $\Omega$ as $n \rightarrow \infty$ with probability one.   We also obtain convergence rates. The proof of Theorem \ref{thm:C} is provided in Appendix \ref{sec:appendixD}.

\begin{theorem}
\label{thm:C}
Assume $s_n = \sigma^2 + c n^{-1/2} \sqrt{\log n}$~for all $n \geq 1$, where $c > 8 (K^2 + \sigma_0^2)$ is  a constant. 

\begin{enumerate}
\item[(a)]
Under A1--A5,  we have 
\[\sup_{\boldsymbol{x} \in \Omega} |\hat{f}_n^{***} (\boldsymbol{x}) - f_0(\boldsymbol{x})| \rightarrow 0\]
as $n \rightarrow \infty$ with probability one.
\item[(b)]
Under A1--A6, we have 
\[\sup_{\boldsymbol{x} \in \Omega} \sup_{\boldsymbol{\beta} \in \partial \hat{f}_n^{***} (\boldsymbol{x})} |\boldsymbol{\beta} - \nabla f_0(\boldsymbol{x})| \rightarrow 0\]
as $n \rightarrow \infty$ with probability one.
\item[(c)]
Under A1--A5, we have 
\begin{equation*}
\frac{1}{n}\sum_{i = 1}^n (\hat{f}_n^{***}(\boldsymbol{X}_i) - f_0(\boldsymbol{X}_i))^2 = \left\{
\begin{array}{l l}
\mathcal{O}_p(n^{-1/2}\sqrt{\log n}), & \mbox{ if } d = 1, 2, 3\\
\mathcal{O}_p(n^{-1/2} \log n), & \mbox{ if } d = 4\\
\mathcal{O}_p(n^{-2/d}), & \mbox{ if } d \geq 5
\end{array}
\right.
\end{equation*}
as $n \rightarrow \infty$.
\end{enumerate}
\end{theorem}

\begin{remark}
\label{rmk:02} It is important to note that the convergence rate of $\hat{f}_n^{***}$ is slower than that of $\hat{f}^*_n$ or $\hat{f}_n^{**}$ when $d = 1, 2$, or $3$. This slow rate is obtained because we set $s_n = \sigma^2 + e_n$, where $e_n = cn^{-1/2}\sqrt{\log n}$. This assumption was made to ensure that $f_0$ is a feasible solution to Problem (C). If a faster rate for $e_n$ could be used to guarantee that $f_0$ remains feasible to Problem (C), we could achieve faster convergence rates for $\hat{f}_n^{***}$ when $d = 1, 2$, or $3$. This is an area that warrants further investigation.
\end{remark}

\section{Numerical Results}
\label{label:numerical_results}
This section is devoted to the investigation of the numerical performance of $\hat{f}_n^*$, $\hat{f}_n^{**}$, and $\hat{f}_n^{***}$ when $J(f) = J_{\infty}(f)$ in the case where $f_0(\boldsymbol{x})$  represents the steady--state waiting time in a single--server queue with a unit arrival rate and a service rate of $\boldsymbol{x} \in [1.2, 1.3]$.

Each of Problems (A), (B), and (C) involves specific parameters; Problem (A) uses $\lambda_n$, Problem (B) uses $u_n$, and Problem (C) uses $s_n$. The selection of these parameters is crucial because it significantly influences the numerical performance of the solutions to Problems (A), (B), and (C). Below, we outline the selection process of these parameters  in our numerical example in Section \ref{num:mm1}.

\paragraph{Parameter $\lambda_n$ for Problem (A)}
We employ a 5-fold cross-validation approach to determine $\lambda_n$. We first divide the data set $D = ((\boldsymbol{X}_i, y_i) : i = 1, \cdots, n)$ into five equally sized subsets, say $D_1, D_2, D_3, D_4,$ and $D_5$. For each $\lambda \geq 0$, we define the following cross-validation function:
\[\mbox{CV}(\lambda) \triangleq \frac{1}{5} \sum_{k = 1}^5 \sum_{(\boldsymbol{X}_i, y_i) \in D_k} (y_i - f_{\lambda, n}^{[k]}(\boldsymbol{X}_i))^2,\]
where $f_{\lambda, n}^{[k]}$ represents the solution to Problem (A) obtained with $\lambda_n = \lambda$, using the data points $(\boldsymbol{X}_i, y_i)$ except for those in $D_k$. We then compute $\mbox{CV}(\lambda)$ for each candidate value $\lambda$ in the set  $\{\lambda^1, \cdots, \lambda^m\}$ and select the value that minimizes $\mbox{CV}(\lambda)$. 

\paragraph{Parameter $u_n$ for Problem (B)} The value of $u_n$ should be selected to ensure $J_{\infty}(f_0) \leq u_n$. Ideally, $u_n$ is the smallest value that satisfies this condition, meaning $u_n = J_{\infty}(f_0)$, if possible. Since we cannot compute $f_0$ exactly, we will use $\hat{f}_n^{***}$ as an estimate of $f_0$. Therefore, after calculating $\hat{f}_n^{***}$ from (\ref{ProblemC_QP}) and (\ref{eqn:05}), we use  $J_{\infty}(\hat{f}_n^{***})$ as an estimate of $u_n$ and proceed to solve  (\ref{ProblemB_QP}) and (\ref{eqn:04}) to determine $\hat{f}_n^{**}$.

\paragraph{Parameter $s_n$ for Problem (C)} To estimate $s_n$ by computing $\mathbb{E}[\mbox{var}(\varepsilon_1|\boldsymbol{X}_1)]$, we partition $[a, b]^d$ into $r$ hyperrectangles, say $H_1, \cdots, H_r$. For each $j \in \{1, \cdots, r\}$, we compute the sample variance $V_j$ of the $y_i$ values corresponding to the $\boldsymbol{X}_i$ values that fall within $H_j$. We then compute $\mathbb{P}(\boldsymbol{X}_1 \in H_1)\cdot V_1 + \cdots + \mathbb{P} (\boldsymbol{X}_1 \in H_r) \cdot V_r$ as an estimate of $s_n$. When the $\boldsymbol{X}_i$ values are uniformly distributed or evenly spaced, the simple average of the $V_j$ values can be used as an estimate of $s_n$.

\subsection{M/M/1 queue}

\label{num:mm1}
We considered the case where $f_0(\boldsymbol{x})$  is the steady--state waiting time in an M/M/1 queue operating under a first in/first out discipline, with a unit arrival rate, and a service rate $\boldsymbol{x}$ in the range $[1.2, 1.3]$. This model was selected because it has a closed-form formula for $f_0$, specifically $f_0 = 1/(\boldsymbol{x}(\boldsymbol{x}-1))$ for $\boldsymbol{x} >1 $, allowing for easy comparison of our estimators to the true values. We set $\boldsymbol{X}_i = 1.2 + 0.1(i-1)/n + 0.1/(2n)$ for $i = 1, \cdots, n$. Next, we simulated the M/M/1 queue initialized empty and idle with the service rate set to $\boldsymbol{X}_i$. The value $y_i$ was taken to be the average waiting time of the first 5000 customers for each $i \in \{1, \cdots, n\}$. Once we had the pairs $(\boldsymbol{X}_i, y_i)$, we computed the estimators  $\hat{f}_n^*$ from (\ref{ProblemA_QP}) and (\ref{eqn:03}), $\hat{f}_n^{**}$ from (\ref{ProblemB_QP}) and (\ref{eqn:04}), and $\hat{f}_n^{***}$ from (\ref{ProblemC_QP}) and (\ref{eqn:05}), by using CVX  \citep{GrantBoyd2014} and MOSEK.

To determine  $s_n$, we partitioned $[1.2, 1.3]$ into $r$ subintervals of the same length, computed the sample variance of the $y_i$'s corresponding to the $\boldsymbol{X}_i$ values that fall within each subinterval, and took the simple average of these sample variances as an estimate of $s_n$. We used  $r = 8, 10$, and $16$ for  $n = 120, 300$, and $400$, respectively. After solving Problem (C) with $s_n$, we solved Problem (B) using $J_{\infty}(\hat{f}_n^{***})$ as $u_n$. We  next solved Problem (A) using two different sets of $(\lambda_n : n \geq 1)$. We first set  $\lambda_n(1) = 1/n^{0.8}$ for $n \geq 1$. Next, we selected $\lambda_n(2)$ through five-fold cross validation on Problem (A), minimizing $\mbox{CV}(\lambda)$ over the set of potential $\lambda$ values $\{0, 10^{-10}, 10^{-6}, 10^{-4}, 10^{-2}, 1, 10^2, 10^6\}$.

\begin{figure}
\caption{(a) The circles are the $(\boldsymbol{X}_i, y_i)$ values; the solid line is $f_0$; the dotted line is $\hat{f}_n^*$, a solution to Problem (A); and the dashed line is $\hat{f}_n^{***}$, a solution to Problem (C). (b) The solid line is $\nabla f_0$; the dotted line is $\mbox{subgrad } \hat{f}_n^*$; and the dashed line is $\mbox{subgrad }\hat{f}_n^{***}$.}
\begin{center}
\includegraphics[scale = 0.3]{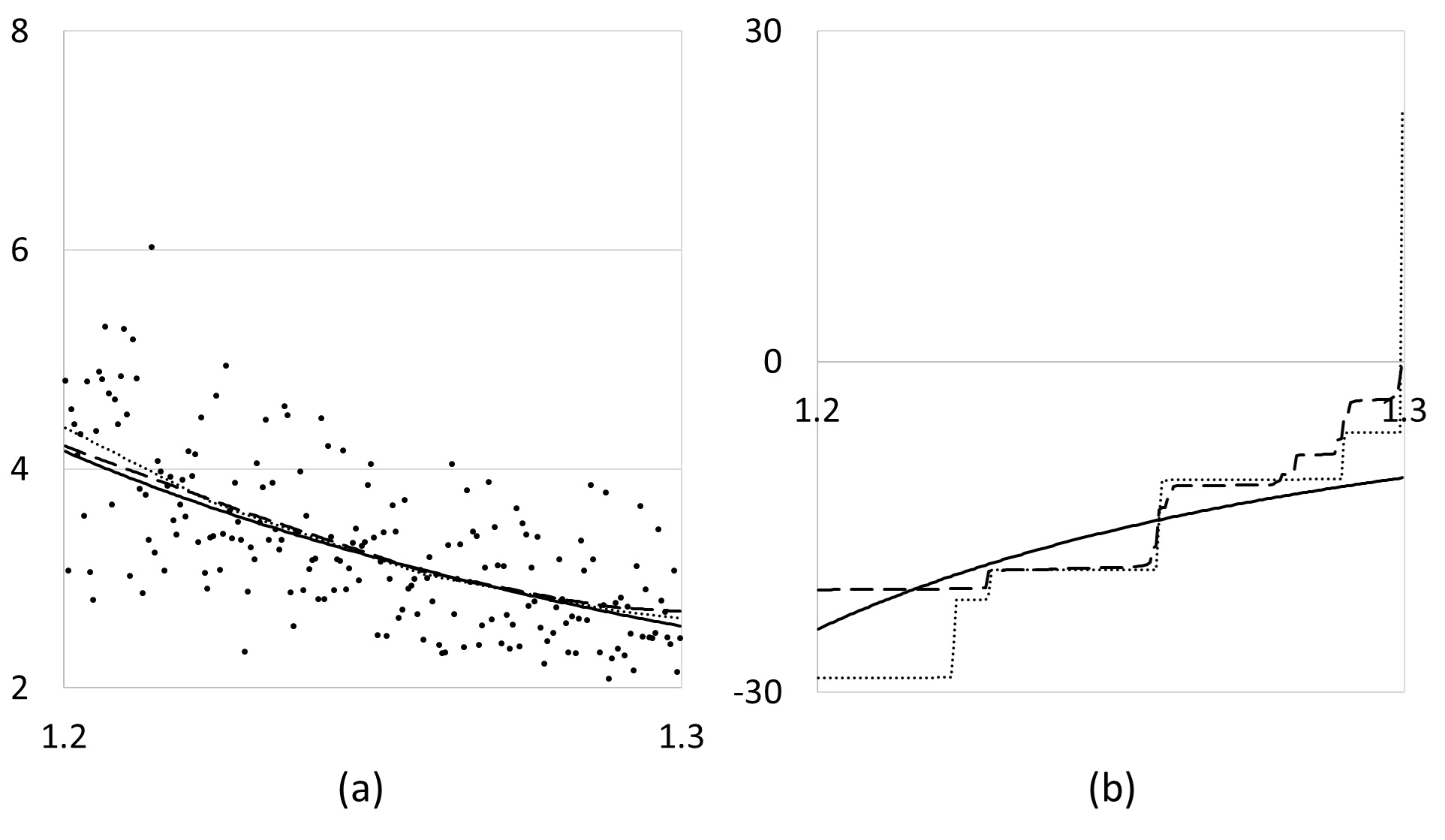}
\end{center}
\label{figure07}
\end{figure}

Figure \ref{figure07} (a) displays the graphs of $\hat{f}_n^*$, a solution to Problem (A), and $\hat{f}_n^{***}$, a solution to Problem (C). Figure \ref{figure07} (b) shows the graphs of their subgradients, revealing that the subgradients from  Problem (C) are consistently closer to the true values of $\nabla f_0$, compared to those from Problem (A).

\begin{figure}
\caption{Box plots for Problem (A) using $\lambda_n(1)$, Problem (A) using $\lambda_n(2)$, Problem (B), and Problem (C) are shown, each based on 100 iid copies of the MAE with respect to $f_0$ from the M/M/1 example.}
\begin{center}
\includegraphics[scale = 0.35]{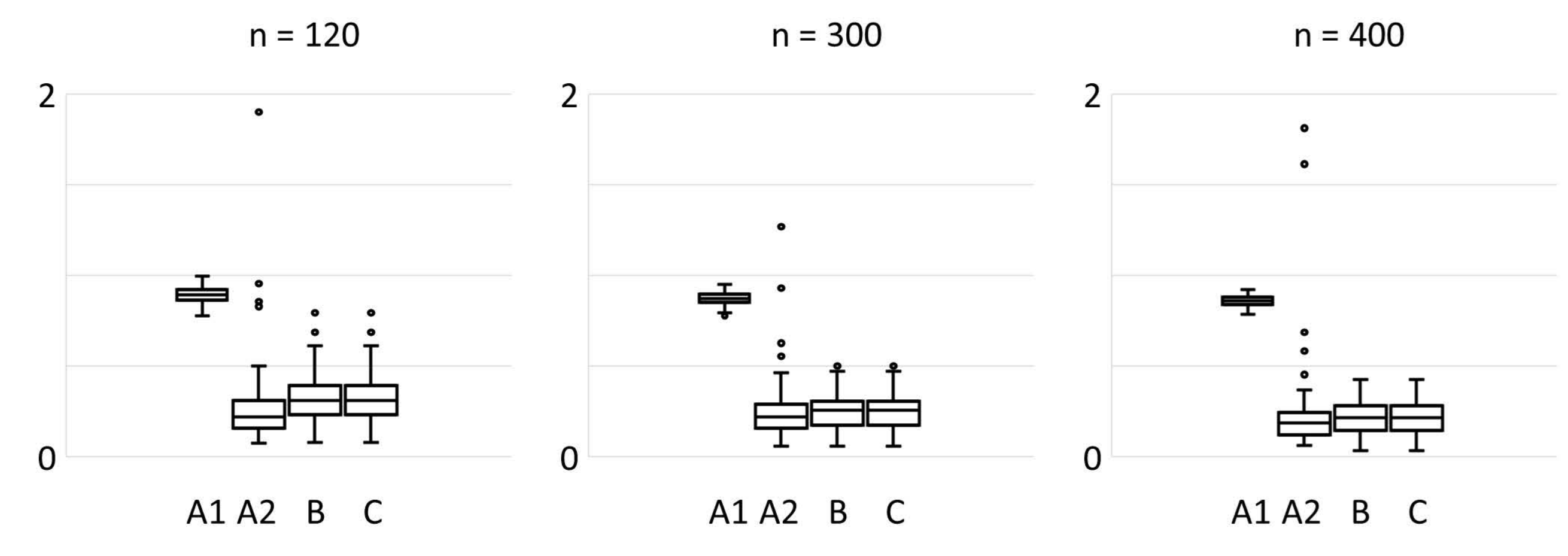}
\end{center}
\label{figure05}

\caption{Box plots for Problem (A) using $\lambda_n(1)$, Problem (A) using $\lambda_n(2)$, Problem (B), and Problem (C) are shown, each based on 100 iid copies of the MAE with respect to $\nabla f_0$ from the M/M/1 example.}
\begin{center}
\includegraphics[scale = 0.35]{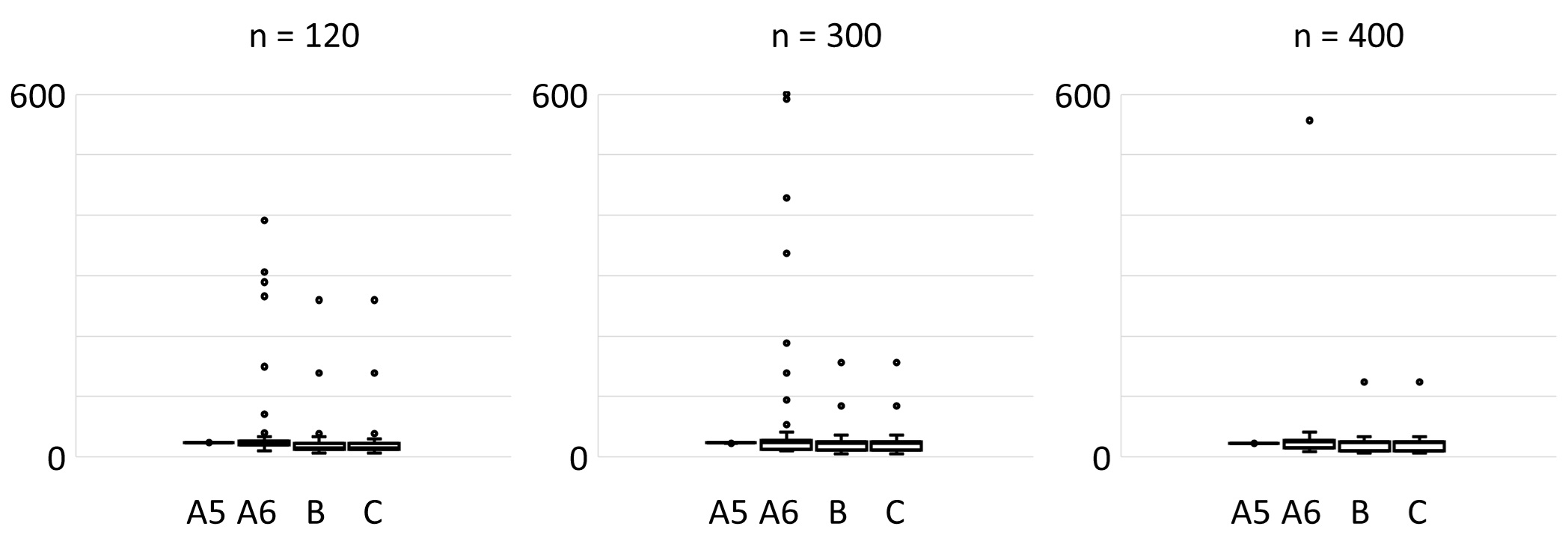}
\end{center}
\label{figure06}
\end{figure}

To evaluate the performance of the three estimators, $\hat{f}_n^*$, $\hat{f}_n^{**}$, and $\hat{f}_n^{***}$, across the entire domain $\Omega = [-1, 1]$, we computed the maximum absolute error (MAE) of  $\hat{f}_n^*$ with respect to $f_0$ as follows:
\begin{equation}
    \label{max_abs_value}
\max_{1 \leq i \leq n} |\hat{f}_n^*(\boldsymbol{X}_i)  - f_0(\boldsymbol{X}_i)|
\end{equation}
and the MAE of $\mbox{subgrad }\hat{f}_n^*$ with respect to $\nabla f_0$ as follows:
\begin{equation} 
    \label{max_abs_derivative}
\max_{1 \leq i \leq n} |\mbox{subgrad }\hat{f}_n^* (\boldsymbol{X}_i)  - \nabla f_0(\boldsymbol{X}_i)|.
\end{equation}
We performed this process independently 100 times, generating 100 replications of each of (\ref{max_abs_value}) and (\ref{max_abs_derivative}). Figure \ref{figure05} shows the box plots created with these 100 iid copies of (\ref{max_abs_value}) for Problem (A) when $\lambda_n(1)$ is used for the smoothing constant. They are displayed above A1. We repeated this process for Problem (A) when $\lambda_n(2)$ was used for the smoothing constant. We also repeated this process for Problem (B) and Problem (C). The box plots corresponding to these case are displayed above A2, B, and C, respectively, in Figure \ref{figure05}. Each box plot shows the median (indicated by the center line), the first quantile (lower edge), the third quartile (upper edge), and outliers represented as small circles. 

Figure \ref{figure06} reports the box plots created with 100 iid copies of (\ref{max_abs_derivative}) when $\lambda_n(1)$ is used for $(\lambda_n: n \geq 1)$. They are displayed above A1. We repeated this process for Problem (A) when $\lambda_n(2)$ was used for the smoothing constant. We also repeated this process for Problem (B) and Problem (C). The box plots corresponding to these case are displayed above A2, B, and C, respectively, in Figure \ref{figure06}. 

Finally, Tables  \ref{mm1_original} and \ref{mm1_gradient} report the 95\% confidence intervals computed from the 100 iid values of the MAD for each case of A1, A2, B, and C.

\begin{table}[ht]
\centering
\caption{95\% confidence intervals of the expected MAE with respect to $f_0$  in the M/M/1 example}
\begin{tabular}{c c c c c c c}
\hline
$n$ & A1 & A2 & B & C\\
\midrule
$120$ & $0.890\pm0.010$ & $0.276\pm0.048$ & $0.322\pm0.030$ & $0.322\pm0.030$\\
$300$ & $0.872\pm0.007$ & $0.253\pm0.036$ & $0.243\pm0.020$ & $0.243\pm0.020$\\     
$400$ & $0.858\pm0.006$ & $0.232\pm0.054$ & $0.213\pm0.018$ & $0.213\pm0.018$\\
\hline
\end{tabular}
\label{mm1_original}

\caption{95\% confidence intervals of the expected MAE with respect to $\nabla f_0$ in the M/M/1 example}
\begin{tabular}{c c c c c c c}
\hline
$n$ & A1 & A2 & B & C\\
\midrule
$120$ & $23.62\pm0.02$ & $75.72\pm54.09$ & $21.21\pm6.17$ & $21.04\pm6.17$\\
$300$ & $23.12\pm0.02$ & $76.09\pm45.09$ & $21.18\pm3.76$ & $21.17\pm3.76$\\
$400$ & $22.83\pm0.02$ & $92.79\pm66.58$ & $20.32\pm3.05$ & $20.31\pm3.05$\\
\hline
\end{tabular}
\label{mm1_gradient}
\end{table}

\subsection{Observations from Numerical Experiments}

The box plots in Figures \ref{figure05} and \ref{figure06} indicate that Problem (A) occasionally generated prohibitively large outliers when $\lambda_n$ was selected via cross validation, which is undesirable.  In contrast, when $\lambda_n$  was chosen as an arbitrary sequence converging to 0 as $n \rightarrow \infty$ in the scenario of A1, both the solution to Problem (A) and its subgradient converged to the true values in Tables \ref{mm1_original} and \ref{mm1_gradient}. However, there was no clear guideline for selecting $\lambda_n$ in this manner, and these arbitrarily chosen $ \lambda_n$ values led to  worse performance compared to those of Problems (B) and (C) in Tables \ref{mm1_original} and \ref{mm1_gradient}. On the contrary, the solution to Problem (C) and its subgradient yielded consistently well-behaved estimators with decreasing MADs as $n$ increased.  Overall, Problem (C) exhibited superior performance compared to Problem (A) in Tables \ref{mm1_original} and \ref{mm1_gradient}. The performance of Problem (B) was comparable to that of Problem (C) because its parameter, $u_n$, was derived from Problem (C).

Figure \ref{figure07} demonstrates a case in which Problem (C) produces estimates of $\nabla f_0$  that are closer to $\nabla f_0$ compared to Problem (A) across the entire domain $\Omega$.
  
\section{Conclusions}
\label{label:conclusions}

In this paper, we introduced a new method for estimating a convex function $f_0$. The new estimator is designed to mitigate the overfitting tendencies of the convex regression estimator by minimizing a penalty term $J(f)$ on the subgradient, over the set of convex functions $f$, while enforcing an upper bound on the sum of squared errors, i.e., $\frac{1}{n}\sum_{i = 1}^n (y_i - f(\boldsymbol{X}_i))^2 \leq s_n$.  Both theoretical and empirical studies show that the proposed estimator exhibits desired properties, including a.s.\ uniform consistency for the estimator and its subgradient over the entire domain.

We conclude this section by highlighting several potential future research topics. 
\begin{enumerate}
\item[(1)]
This paper specifically addresses the case where $J(f) = J_{\infty}(f)$. An interesting question for future research is how estimators behave when $J(f) = J_p (f)$ for $1 \leq p < \infty$ and whether they are computable as solutions to quadratic programs.

\item[(2)]
Another valuable question for research is whether we can improve the convergence rate for Problem (C) by using an improved estimate of $s_n$; further details can be found in Remark \ref{rmk:01}. 
\item[(3)]
Examining the relationship between Problems (B) and (C) is also a promising research direction.
\item[(4)] In this paper, we focused on the case where $\mathcal{F}$ is the set of convex functions with finite $J(f)$. However, the formulations presented in Problems (A), (B), and (C) can be adapted for various  other shape--restricted regression problems. For example, $\mathcal{F}$ could be the set of monotone functions with finite $J(f)$. Further investigation of these variations is  a promising research area.
\end{enumerate}



%
%
%
\begin{appendices}
\section{Proof of Proposition \ref{proposition:01}}
\label{sec:appendixA}


Note that (\ref{ProblemA_QP}) is a minimization problem of a continuous and coersive function over a closed subset in $\mathbb{R}^{n(d + 1) + 1}$, so its minimum point $\hat{f}_1, \cdots, \hat{f}$, $\hat{\boldsymbol{\beta}}_1, \cdots, \hat{\boldsymbol{\beta}}_n$, $\hat{M}$ exists (Proposition 7.3.1 and Theorem 7.3.7 of \cite{KurdilaZabarankin2006}). Since the objective function in (\ref{ProblemA_QP}) is strictly convex in $f_1, \cdots, f_n$ and $M$, the values $\hat{f}_1, \cdots, \hat{f}_n$ and $\hat{M}$ are unique.

We now need to show that $\hat{f}_n^*$ is a solution to Problem (A) with $J(f) = J_{\infty}(f)$. Note that $\hat{f}_n^*$ is a convex function and satisfies $\hat{M} \geq J_{\infty} (\hat{f}_n^*)$, which implies $\hat{f}_n^* \in \mathcal{F}$. For any convex function $g \in \mathcal{F}$, we can define $l_g:\Omega \rightarrow \mathbb{R}$ by $l_g(\boldsymbol{x}) = \max_{1 \leq i \leq n} \{g(\boldsymbol{X}_i) + \boldsymbol{\gamma}_i^{\mathsf{T}} (\boldsymbol{x} - \boldsymbol{X}_i)\}$, where $\boldsymbol{\gamma}_i$ is a subgradient of $g$ at $\boldsymbol{X}_i$ such that there exist sequences $\boldsymbol{z}_1, \boldsymbol{z}_2, \cdots \in \Omega_g$ tending to $\boldsymbol{X}_i$ and $\nabla g(\boldsymbol{z}_1), \nabla g(\boldsymbol{z}_2), \cdots$ tending to $\boldsymbol{\gamma}_i$. Such sequences always exist due to Theorem 25.5 on page 246 and Theorem 24.4 on page 233 of \cite{Rockafella}. It should be noted that $g(\boldsymbol{X}_1), \cdots, g(\boldsymbol{X}_n)$, $\boldsymbol{\gamma}_1, \cdots, \boldsymbol{\gamma}_n$, and $ \max_{1 \leq i \leq n} |\boldsymbol{\gamma}_i|_{\infty}$ form a feasible solution to (\ref{ProblemA_QP}). Therefore,
\begin{align*}
\frac{1}{n}\sum_{i = 1}^n (Y_i - \hat{f}_n^*(\boldsymbol{X}_i))^2 + \lambda_n J_{\infty}(\hat{f}_n^*) & \leq \frac{1}{n}\sum_{i = 1}^n (Y_i - \hat{f}_i)^2 + \lambda_n \hat{M}\\
&\leq \frac{1}{n}\sum_{i = 1}^n (Y_i - g(\boldsymbol{X}_i))^2 + \lambda_n \max_{1 \leq i \leq n }|\boldsymbol{\gamma}_i|_{\infty}\\
&\leq   \frac{1}{n}\sum_{i = 1}^n (Y_i - g(\boldsymbol{X}_i))^2 + \lambda_n J_{\infty} (g).
\end{align*}
Hence, $\hat{f}_n^*$ is a solution to Problem (A).

\section{Proof of Proposition \ref{proposition:02}}
\label{sec:appendixB}

Since the $\varepsilon_i$'s are uniformly sub-Gaussian, the $\varepsilon_i^2$'s satisfy the Bernstein condition. According to  Bernstein's inequality (Lemma 8.2 on page 127 of \cite{vanDeGeer2000}), we have
\begin{equation}
\label{eqn:10} \mathbb{P}\left(\frac{1}{n}\sum_{i = 1}^n (y_i - f_0(\boldsymbol{X}_i))^2 > s_n \right)=  \mathbb{P}\left(\frac{1}{n}\sum_{i = 1}^n (\varepsilon_i^2 - \sigma^2) > c n^{-1/2} \sqrt{\log n} \right)  \leq n^{-c/(8(K^2 + \sigma_0^2))}.
\end{equation}
Thus, $\sum_{n = 1}^{\infty} \mathbb{P}\left(\frac{1}{n}\sum_{i = 1}^n (y_i  - f_0(\boldsymbol{X}_i))^2  > s_n \right) < \infty$
and the Borel-Cantelli lemma implies $\frac{1}{n}\sum_{i= 1}^n  (y_i- f_0(\boldsymbol{X}_i))^2 \leq s_n$ for all $n$ sufficiently large a.s. The remainder of Proposition \ref{proposition:03} follows from  arguments similar to those in the proof of Proposition \ref{proposition:01}.

\section{Proof of Theorem \ref{thm:A}}
\label{sec:appendixC}


The proof of Theorem \ref{thm:A} consists of the following steps.

\textit{\textbf{Step 1:}} Let $\hat{f}_1, \cdots, \hat{f}_n, \hat{\boldsymbol{\beta}}_1, \cdots, \hat{\boldsymbol{\beta}}_n, \hat{M}$ be a solution to (\ref{ProblemA_QP}). We first establish 
\begin{equation}
\label{eqn:06} J_{\infty} (\hat{f}_n^*) = \max_{1 \leq i \leq n} |\hat{\boldsymbol{\beta}}_i|_{\infty}.
\end{equation}
To prove this by contradiction, suppose, on the contrary, $J_{\infty} (\hat{f}_n^*) < \max_{1 \leq i \leq n} |\hat{\boldsymbol{\beta}}_i|_{\infty}$. This implies that there exists $r \in \{1, \cdots, n\}$ such that $\max_{1 \leq i \leq n}|\hat{\boldsymbol{\beta}}_i| = \hat{\boldsymbol{\beta}}_r$, $\hat{f}_n^* (\boldsymbol{x}) = \hat{f}_r + \hat{\boldsymbol{\beta}}_r^{\mathsf{T}}(\boldsymbol{x} - \boldsymbol{x}_r)$ on a set of measure zero, and $\hat{f}_n^*$ is not differentiable at $\boldsymbol{x}_r$. 

By modifying $\hat{\boldsymbol{\beta}}_r$, we can construct another minimizer of (\ref{ProblemA_QP}). Define $\hat{g}_n^*:\Omega \rightarrow \mathbb{R}$ by 
\[\hat{g}_n^* (\boldsymbol{x}) = \max_{1 \leq i \leq n, i \neq r} \{\hat{f}_i + \hat{\boldsymbol{\beta}}_i^{\mathsf{T}}(\boldsymbol{x} - \boldsymbol{X}_i)\}\]
for $\boldsymbol{x} \in \Omega$. Then, for $\boldsymbol{x} \in \Omega$, $\hat{f}_n^* (\boldsymbol{x}) = \hat{g}_n^*(\boldsymbol{x})$. According to Theorem 24.4 on page 233 of \cite{Rockafella}, there exist sequences $\boldsymbol{z}_1, \boldsymbol{z}_2, \cdots \in \Omega_{\hat{f}_n^*}$ tending to $\boldsymbol{x}_r$ and $\nabla \hat{g}_n^* (\boldsymbol{z}_1) = \nabla \hat{g}_n^* (\boldsymbol{z}_2) = \cdots = \hat{\boldsymbol{\beta}}_j$ for some $j \in \{1, \cdots, n\} \setminus \{r\}$. By Theorem 25.5 on page 246 of \cite{Rockafella}, $\hat{\boldsymbol{\beta}}_j$ is a subgradient of $\hat{g}_n^*$ at $\boldsymbol{x}_r$. 

Thus, if we take $\hat{f}_1, \cdots, \hat{f}_n, \hat{\boldsymbol{\beta}}_1, \cdots, \hat{\boldsymbol{\beta}}_n, \hat{M}$, replace $\hat{\boldsymbol{\beta}}_r$ by $\hat{\boldsymbol{\beta}}_j$, and replace $\hat{M}$ by $\max_{1 \leq i \leq n, i \neq r} |\hat{\boldsymbol{\beta}}_i|_{\infty}$, we create a new minimizer of (\ref{ProblemA_QP}) whose objective value is less than $\frac{1}{n}\sum_{i = 1}^n (y_i - \hat{f}_i)^2 + \lambda_n \hat{M}$, which leads to a contradiction.

\textit{\textbf{Step 2:}} We next establish that there exists a constant $\alpha$ such that 
\begin{equation}
\label{eqn:07} \sup_{\boldsymbol{x} \in \Omega} \sup_{\boldsymbol{\beta} \in \partial \hat{f}_n^* (\boldsymbol{x})} |\boldsymbol{\beta}|_{\infty} \leq \alpha
\end{equation}
and
\begin{equation}
\label{eqn:08}
\frac{1}{n}\sum_{i = 1}^n (\hat{f}_n^*(\boldsymbol{X}_i) - f_0(\boldsymbol{X}_i))^2 \leq \alpha \lambda_n + 2\delta_n^2
\end{equation}
for $n$ sufficiently large a.s.

To establish (\ref{eqn:07}) and (\ref{eqn:08}), we need some definitions. Let 
\begin{align*}
&\mathcal{G} = \{f:\Omega \rightarrow \mathbb{R} \mid \mbox{ There exist }\alpha_1, \cdots, \alpha_n \in \mathbb{R} \\
&\mbox{ and } \boldsymbol{\beta}_1, \cdots, \boldsymbol{\beta}_n \in \mathbb{R}^d \mbox{ satisfying }\\
& f(\boldsymbol{x}) = \max_{1 \leq i \leq n} \{\alpha_i + \boldsymbol{\beta}_i^{\mathsf{T}}(\boldsymbol{x} - \boldsymbol{X}_i)\} \mbox{ for } \boldsymbol{x} \in \Omega,\\
& J_{\infty} (f) = \max_{1 \leq i \leq n} | \boldsymbol{\beta}_i|_{\infty}, \mbox{ and }\\
& |f(1/2, \cdots, 1/2) - f_0(1/2, \cdots, 1/2))| \leq 1\}.
\end{align*}
Then $\hat{f}_n^* \in \mathcal{G}$ for all $n$ sufficiently large a.s.\ as indicated by  (\ref{eqn:06}) and arguments similar to those in the proof of Theorem 1 of \cite{Lim2021}.  Next, define 
\begin{equation*}
\mathcal{G}_*(\delta) = \biggl\{f \in \mathcal{G} \mid \frac{1}{n}\sum_{i = 1}^n (f(\boldsymbol{X}_i) - f_0(\boldsymbol{X}_i))^2 + \lambda_n J_{\infty} (f) \leq \delta^2\biggr\}
\end{equation*}
for all $\delta > 0$. Then the $u$-entropy of $\mathcal{G}_*(\delta_n)$ in the supremum norm is of order $u^{-d/2}$ because $\mathcal{G}_*(\delta_n)$ is a subset of 
\begin{equation}
\label{eqn:11}\mathcal{G}' = \Bigl\{f:\Omega \rightarrow \mathbb{R} \mid  f \mbox{ is convex,}\; |f(\boldsymbol{x})| \leq c_1 
\mbox{ for all } \boldsymbol{x} \in \Omega, \mbox{ and } \sup_{\boldsymbol{x} \in \Omega} \sup_{\boldsymbol{\beta} \in \partial f(\boldsymbol{x})} |\boldsymbol{\beta}|_{\infty} \leq c_2\Bigr\}
\end{equation}
for some constants $c_1$ and $c_2$,  we have $\delta_n^2/\lambda_n = O(1)$, and the $u$-entropy of $\mathcal{G}'$ in the supremum norm is of order $u^{-d/2}$ by Theorem 6 of \cite{Bronshtein1976}.

Now, we can apply Theorem 2.1 of \cite{vanDeGeer2001} to conclude
\begin{equation}
\label{eqn:09}\mathbb{P}\biggl( \frac{1}{n}\sum_{i = 1}^n (\hat{f}_n^*(\boldsymbol{X}_i) - f_0(\boldsymbol{X}_i))^2 + \lambda_n J_{\infty}(\hat{f}_n^*) \geq 
2 \lambda_n J_{\infty} (f_0) + 2 \delta_n^2\biggr) \leq c_3 \exp (-n \delta_n^2/c_3)
\end{equation}
for some constant $c_3 >0$.

By (\ref{eqn:09}) and the Borel-Cantelli lemma, we have 
\begin{equation}
\label{eqn:09a}\frac{1}{n}\sum_{i = 1}^n (\hat{f}_n^*(\boldsymbol{X}_i) - f_0(\boldsymbol{X}_i))^2 + \lambda_n J_{\infty}(\hat{f}_n^*)
 \leq 2 \lambda_n J_{\infty} (f_0) + 2 \delta_n^2\end{equation}
for $n$ sufficiently large a.s., thereby implying (\ref{eqn:08}). Since $\delta_n^2 = O(\lambda_n)$, (\ref{eqn:09a}) implies (\ref{eqn:07}).

\textit{\textbf{Step 3:}} We will now use (\ref{eqn:07}), (\ref{eqn:08}), and the fact that $f_0 \in \mathcal{F}$ to establish Theorem 1 (a). Let $\epsilon > 0$ be given. Since $\Omega$ is compact, we can find a finite collection of sets $S_1, \cdots, S_l$ covering $\Omega$, each having a diameter less than $\epsilon$, i.e., $|\boldsymbol{x} - \boldsymbol{y}|_2 \leq \epsilon$ for any $\boldsymbol{x}, \boldsymbol{y} \in S_i$, $i = 1, \cdots, l$. By (\ref{eqn:07}) and the fact that $f_0 \in \mathcal{F}$, we can find a Lipschitz constant, say $\gamma$, for both $\hat{f}_n^*$ and $f_0$ over $\Omega$. For each $\boldsymbol{x} \in S_j$ and $\boldsymbol{X}_i \in S_j$,
\begin{align*}
|\hat{f}_n^* (\boldsymbol{x}) - f_0(\boldsymbol{x})|  &\leq |\hat{f}_n^*(\boldsymbol{x}) - \hat{f}_n^*(\boldsymbol{X}_i)|
 + |\hat{f}_n^*(\boldsymbol{X}_i) - f_0(\boldsymbol{X}_i)| + |f_0(\boldsymbol{X}_i) - f_0(\boldsymbol{x})|\\
 &\leq  \gamma \epsilon + |\hat{f}_n^*(\boldsymbol{X}_i) - f_0(\boldsymbol{X}_i)| + \gamma \epsilon.
\end{align*}
So, 
\begin{align*}
\sup_{\boldsymbol{x} \in S_j} |\hat{f}_n^*(\boldsymbol{x}) - f_0(\boldsymbol{x})| & \leq 2\gamma\epsilon + \frac{\sum_{i = 1}^n |\hat{f}_n^*(\boldsymbol{X}_i) - f_0(\boldsymbol{X}_i)| I(\boldsymbol{X}_i \in S_j)}{\sum_{i = 1}^n I(\boldsymbol{X}_i \in S_j)}\\
 & \leq 2\gamma\epsilon + \frac{1}{n}\sum_{i = 1}^n |\hat{f}_n^*(\boldsymbol{X}_i) - f_0(\boldsymbol{X}_i)| \cdot \frac{n}{\sum_{i = 1}^n I(\boldsymbol{X}_i \in S_j)}\\
 & \leq 2 \gamma \epsilon +\sqrt{\frac{1}{n}\sum_{i = 1}^n (\hat{f}_n^*(\boldsymbol{X}_i) - f_0(\boldsymbol{X}_i))^2}  \cdot\frac{n}{\sum_{i = 1}^n I(\boldsymbol{X}_i \in S_j)}.
\end{align*}
From (\ref{eqn:08}) and the fact that $\lambda_n \rightarrow 0$ and $\delta_n^2 \rightarrow 0$ as $n \rightarrow \infty$, we conclude 
\[\limsup_{n \rightarrow \infty} \sup_{\boldsymbol{x} \in S_j} |\hat{f}_n^* (\boldsymbol{x}) - f_0(\boldsymbol{x})| \leq 2\gamma \epsilon\]
a.s.\ Because there are finitely many $S_j$'s, we can conclude 
\[\sup_{\boldsymbol{x} \in \Omega} |\hat{f}_n^*(\boldsymbol{x}) - f_0(\boldsymbol{x})| \rightarrow 0\]
a.s.\ as $n \rightarrow \infty$, proving Theorem 1 (a). 

\textit{\textbf{Step 4:}} We observe that Theorem 1 (b) can be derived using arguments similar to those in the proof of Theorem 2 of \cite{Lim2021}.

\textit{\textbf{Step 5:}} From the fact that $\lambda_n = O(\delta_n^2)$ and (\ref{eqn:08}), we can conclude Theorem 1 (c).

\section{Proof of Theorem \ref{thm:C}}
\label{sec:appendixD}

The proof of Theorem \ref{thm:C} consists of the following steps.

\textit{\textbf{Step 1:}} We show that for any $\boldsymbol{x} \in \Omega$, there exists a subgradient $\boldsymbol{\beta}$ of $f_0$ at $\boldsymbol{x}$ such that $|\boldsymbol{\beta}|_{\infty} \leq J_{\infty}(f_0)$. If $f_0$ is differentiable at $\boldsymbol{x}$, then $\nabla f_0 (\boldsymbol{x})\leq J_{\infty}(f_0)$ by (\ref{eqn:02}). If $f_0$ is not differentiable at $\boldsymbol{x}$, then there are sequences $\boldsymbol{z}_1, \boldsymbol{z}_2, \cdots \in \Omega_{f_0}$ tending to $\boldsymbol{x}$ and $\nabla f_0(\boldsymbol{z}_1), \nabla f_0 (\boldsymbol{z}_2), \cdots$ tending to $\boldsymbol{\xi} \in \partial f_0(\boldsymbol{x})$; see Theorem 25.5 on page 246 and Theorem 24.4 on page 233 of \cite{Rockafella}. Since $|\nabla f_0(\boldsymbol{z}_i)|_{\infty} \leq J_{\infty} (f_0)$ for all $i \geq 1$, it follows that $|\boldsymbol{\xi}|_{\infty} \leq J_{\infty}(f_0)$.

\textit{\textbf{Step 2:}} Let $\hat{f}_1, \cdots, \hat{f}_n, \hat{\boldsymbol{\beta}}_1, \cdots, \hat{\boldsymbol{\beta}}_n, \hat{M}$ be a solution to (\ref{ProblemC_QP}). We establish that $|\hat{\boldsymbol{\beta}}_i|_{\infty} \leq J_{\infty} (f_0)$ for $i  = 1, \cdots, n$ for all $n$ sufficiently large a.s. Specifically,  Step 1 guarantees that there exists $\boldsymbol{\xi}_i \in \partial f_0(\boldsymbol{X}_i)$ with $|\boldsymbol{\xi}_i|_{\infty} \leq J_{\infty}(f_0)$ for all $i \geq 1$. According to Proposition \ref{proposition:03}, $f_0(\boldsymbol{X}_1), \cdots, f_0(\boldsymbol{X}_n)$, $\boldsymbol{\xi}_1, \cdots, \boldsymbol{\xi}_n$, $\max_{1 \leq i \leq n} |\boldsymbol{\xi}_i|_{\infty}$ form a feasible solution to (\ref{ProblemC_QP}) for all $n$ sufficiently large a.s., leading to $|\hat{\boldsymbol{\beta}}_i|_{\infty} \leq J_{\infty} (f_0)$ for $i  = 1, \cdots, n$ for all $n$ sufficiently large a.s.

\textit{\textbf{Step 3:}} Utilizing the fact that the $\varepsilon_i$'s are sub-Gaussian, we establish 
\begin{equation}
\label{eqn:12}\frac{1}{n}\sum_{i = 1}^n (\hat{f}_n^{***} (\boldsymbol{X}_i) - f_0(\boldsymbol{X}_i))^2 \leq \frac{2}{n}\sum_{i = 1}^n \varepsilon_i (\hat{f}_n^{***}(\boldsymbol{X}_i) - f_0(\boldsymbol{X}_i)) + 2cn^{-1/2} \sqrt{\log n}
\end{equation} 
for all $n$ sufficiently large a.s.

Since  $\hat{f}_n^{***}$ satisfies 
\[\frac{1}{n}\sum_{i = 1}^n (y_i - \hat{f}_n^{***}(\boldsymbol{X}_i))^2 \leq s_n\]
for all $n$ sufficiently large a.s., we can derive
\begin{equation*}
\frac{1}{n}\sum_{i = 1}^n (\hat{f}_n^{***}(\boldsymbol{X}_i) - f_0(\boldsymbol{X}_i))^2 \leq \frac{2}{n}\sum_{i = 1}^n \varepsilon_i (\hat{f}_n^{***}(\boldsymbol{X}_i) - f_0(\boldsymbol{X}_i)) - \frac{1}{n}\sum_{i = 1}^n \varepsilon_i^2 + \sigma^2 + cn^{-1/2} \sqrt{\log n}\end{equation*}
for all $n$ sufficiently large a.s.\ Using arguments similar to those leading (\ref{eqn:10}), we can obtain $-\frac{1}{n}\sum_{i = 1}^n \varepsilon_i^2 + \sigma^2 - c n^{-1/2} \sqrt{\log n} \leq 0$ for all $n$ sufficiently large a.s., and hence, reach (\ref{eqn:12}).

\textit{\textbf{Step 4:}} Using Steps 2 and 3, along with arguments similar to those in the proof of Theorem 4 of \cite{LiaoEtAl2024}, we conclude 
\[\frac{1}{n}\sum_{i = 1}^n (\hat{f}_n^{***} (\boldsymbol{X}_i) - f_0(\boldsymbol{X}_i))^2 \rightarrow 0\] as $n \rightarrow \infty$ with probability one. Applying arguments similar to those in the proof of Theorem \ref{thm:A} (a), we also conclude 
\[\sup_{\boldsymbol{x} \in \Omega} |\hat{f}_n^{***}(\boldsymbol{x}) - f_0(\boldsymbol{x})| \rightarrow 0\]
as $n \rightarrow \infty$ with probability one.

\textit{\textbf{Step 5:}} Theorem \ref{thm:C} (b) follows from arguments analogous to those in the proof of Theorem 2 from \cite{Lim2021}.

\textit{\textbf{Step 6:}} By Steps 2 and 4, we see that $\hat{f}_n^{***}$ belongs to $\mathcal{G}'$ defined in (\ref{eqn:11}). Propositions 3 and 4 of \cite{Lim2014} imply
\begin{align}
\label{eqn:13}
&\frac{1}{n}\sum_{i = 1}^n \varepsilon_i (\hat{f}_n^{***} (\boldsymbol{X}_i) - f_0(\boldsymbol{X}_i))\\
&= \left\{ 
\begin{array}{l l}
\displaystyle \mathcal{O}_p \left( n^{-1/2}\left\{ \frac{1}{n}\sum_{i = 1}^n (\hat{f}_n^{***}(\boldsymbol{X}_i) - f_0(\boldsymbol{X}_i))^2\right\}^{\frac{1}{2}(1 - d/4)}\right), & \mbox{ if } d \leq 3\\
\mathcal{O}_p \left(n^{-1/2} \log n \right),  & \mbox{ if } d = 4\\
\mathcal{O}_p (n^{-2/d}),  & \mbox{ if } d \geq 5
\end{array}
\right.\nonumber
\end{align}
as $n \rightarrow \infty$. Combining (\ref{eqn:12}) and (\ref{eqn:13}) yields Theorem \ref{thm:C} (c).

\end{appendices}







\bibliographystyle{apalike}

\end{document}